\documentclass{article}

\usepackage{arxiv}

\usepackage[utf8]{inputenc} 
\usepackage[T1]{fontenc}    
\usepackage{hyperref}       
\usepackage{url}            
\usepackage{booktabs}       
\usepackage{amsfonts}       
\usepackage{nicefrac}       
\usepackage{microtype}      
\usepackage{lipsum}
\usepackage{graphicx}
\usepackage{fancyhdr,graphicx,amsmath,amssymb}
\usepackage[ruled,vlined]{algorithm2e}
\usepackage{palatino}
\graphicspath{ {./images/} }

\title{Distributed Swarm Intelligence}

\author{
 Karthik reddy Kanjula \\
  School of Coumputing and Information\\
  West Chester University of Pennsylvania\\
  West Chester, PA 19383  \\
  \texttt{karthikreddykanjula99@gmail.com} \\
   \And
Sai Meghana Kolla\\
  School of Mathematics and Computer Science\\
  Pennsylvania state University\\
  Harrisburg, PA 17057 \\
  \texttt{szk6163@psu.edu} \\
}

\begin{document}
\include{pythonlisting}
\maketitle
\begin{abstract}
This paper presents the development of a distributed application that facilitates the understanding and application of swarm intelligence in solving optimization problems. The platform comprises a search space of customizable random particles, allowing users to tailor the solution to their specific needs. By leveraging the power of Ray distributed computing, the application can support multiple users simultaneously, offering a flexible and scalable solution. The primary objective of this project is to provide a user-friendly platform that enhances the understanding and practical use of swarm intelligence in problem-solving.
\end{abstract}


\section{Introduction}
The Particle Swarm Optimization (PSO) algorithm is an approximation algorithm that finds the best solution from all the explored feasible solutions for any problem that can be formulated into a mathematical equation. In the field of algorithms and theoretical computer science, optimization problems are known by the name "approximation" algorithms. In this project, we built a web application that hosts a PSO algorithm with interactive features such that any person trying to solve a problem with PSO can leverage our distributed application with Ray to solve it.

\section{Motivation}
\label{sec:headings}
The wide-range availability of models based on neural networks and machine learning algorithms explain future of AI development in today's technology-driven environment. Swarm Intelligence is a branch of AI which is adapted from the nature to solve the problems faced by humans.\\

Swarm Intelligence (S.I.) was first proposed in 1989 by Gerardo Beni and Jing Wang, as the name implies S.I. is collective intelligence. To explain, consider a flock of birds that travel together, every individual bird can make a decision and all the birds in a flock communicate and come up with a decision to migrate to a particular place in a particular pattern depending upon the season.  There are many such examples in our ecosystem that represent Swarm Intelligence like ant colonies, bee colonies, and schools of fish. The basic idea is to bring in a set of agents or particles which have an intelligence of their own and these intelligent systems communicate with each other and reach a common and near-optimal solution for a given problem \cite{gupta}.\\

As mentioned above, the flock of birds inspired developers to develop Particle Swarm Optimization algorithm. In this algorithm, we will have a certain number of particles that will be working together by communicating continuously to achieve a common goal. The applications of PSO in the real world are limitless \cite{Kennedy}.\\

In the next generation of AI applications, the algorithm behaviour is understandable to the end-user when interacting. These interactive applications create new and complex problems like high processing and adaptability. With Ray, a distributed computing framework, new and complex system requirements such as performance and scalability can be addressed. Ray provides a unified interface for expressing task-parallel computation, which is powered by a single dynamic execution engine \cite{Moritz}.\\

The framework we suggested for this project helps in solving problems such as energy storage optimization, NP-hard problems, and others. Any such optimization problem that forms a mathematical equation is solvable by reducing to this algorithm, using our framework makes it a scalable, distributed Python application. The main motivation of our project is to introduce people to what swarm intelligence is and how it can be achieved through PSO by providing them with a visualization of how the algorithm works.\\

\section{Literature survey}

The particle swarm optimization algorithm was first studied by Kennedy and Eberhart (1995) on bird flocking and fish school behavior led to the development of this type of algorithm. The term boids is a contraction of the term birdoid objects and is widely used to denote flocking creatures. Using the social environment concept they described the implement of the particle swarm optimization (PSO) method \cite{lind}.\\

The particle swarm optimization algorithm implemented using python programming language is wrapped with Bokeh for plotting and Panel for dash-boarding. The Panel API offers a high level of flexibility and simplicity. Many of the most popular dashboard functions are provided directly on Panel objects and equally across them, making them easier to deal with. Furthermore, altering a dashboard's individual components, as well as dynamically adding/removing/replacing them, is as simple as manipulating a list or dictionary in Python. A number of basic requirements drove the decision to construct an API on top of Bokeh rather than merely extend it \cite{rud}.\\

The authors in paper \cite{shirako} discussed about a significant issue faced by many domain scientists in figuring out how to design a Python-based application that takes advantage of the parallelism with inherent distributedness and heterogeneous computing. Domain scientists' normal methodology is experimenting with novel methods on tiny datasets before moving on to larger datasets. When the dataset size grows too enormous to be processed on a single node, a tipping point is achieved, similarly  a tipping point can also be reached when accelerators are over utilized.\\

One of the solution to above problem is to use Ray. A worker in a Ray is a stateless process that performs activities (remote functions) which are triggered by a driver or another process. As a process of distributing the application, the system layer in Ray launches workers and assigns them tasks. A computationally intensive task in any algorithm requires distributed solution to optimize performance, such tasks are critically identified and automatically published among workers to solve them practically. A worker tries to solve tasks in a sequential manner, with no local state restrained between them, was explained by \cite{phillipp}. Ray, a distributed framework and the basic Ray core API patterns like remote functions as tasks are used in this project to achieve distributive.\\

\section{Design}\label{ch:2}

System design can easily be put into three components. First, implementation of the algorithm. Second, using bokeh,panel libraries to develop a dashboard for interaction and visualisation of particle swarm optimization algorithm in a client/server approach in an assigned public network for multiple clients. Lastly, the dashboard developed is then integrated with the ray framework to execute code asynchronously while the ray framework takes care of the distribution process. This project implements a distributed web application using ray to achieve distributed computing by parallelizing the code between assigned worker nodes.\\

This project is distributed in three ways:

\begin{enumerate}

\item \textbf{Inherently distributed particle swarm optimization algorithm : }
\newline
As mentioned in the introduction, The Particle Swarm Optimization algorithm is inherently distributed. Each individual particle communicates with one another and comes up with an optimal decision. For instance, if the problem is to find a minimum point where x2 + y2 is minimum, the particles searches the entire search space and each particle lays down the best position that is found and based on all the results, the particles together will come up with the best possible solution.
\\

\item \textbf{Client/Server based dashboard: }
\newline
Using Panel server we are hosting the application online that is available to a system in the same wireless network. Every user that opens the application is a client and the computer on the which the program is running acts a server.\\

\item \textbf{Distributed Computing using Ray: }
\newline
Multiple users accessing the application can increase the load on the computer on which it is running, to overcome this Ray framework is used for distributed computing. 
Ray consists of a head node connected with worker nodes that creates jobs with processes id's and a set of worker nodes including a head node to work on.
\\
\end{enumerate}

\subsection{System Architecture}\

\begin{figure}[ht]
\centering
\includegraphics[width=18cm]{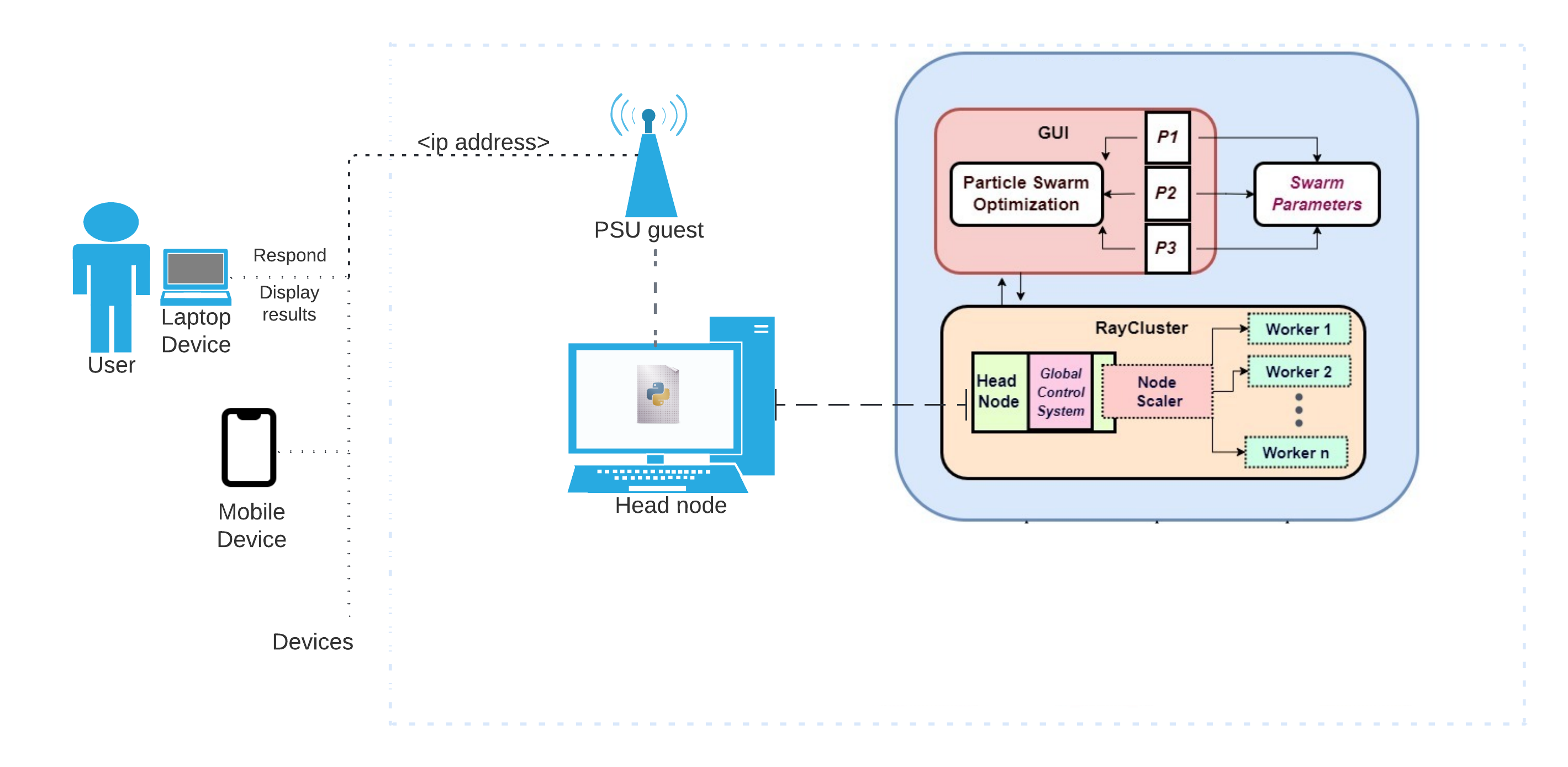}
\caption{Architecture of Distributed Swarm Intelligence}
\end{figure}
\begin{enumerate}

\item \textbf{Clients:} Multiple users or clients can access the application simultaneously and work on the interactive GUI.

\item \textbf{User Interface:} In the interactive UI, a user can individually interact with the particles of a swarm and tweak the parameters and observe the behaviour of the particles.

\item \textbf{Server:} Requests from client are received by the server and the tasks are distributed among worker nodes.\\
\end{enumerate}

\section{Implementation}\label{sec:4}

The project is implemented in Python. It is the best fit for the project because of the access to third party libraries and frameworks for Dashboard and Distributed computing.

\begin{itemize}
\item \textbf{Hardware Heterogeneity} : The application can be accessed from any machine irrespective of the OS.
  
  \item \textbf{Resource Sharing} : Using Ray, multiple computers can be connected together and share resources.
  
  \item \textbf{Concurrency} : Multiple users can connect to the network to access it.
  
  \item \textbf{Scalability} : With ray, any number of worker nodes can be added easily to distribute the computation load.
\end{itemize}

\subsection{Algorithm}

Particle Swarm Optimization algorithm is implemented using python programming language. In Algorithm 1 below, the psuedo code for the PSO algorithm is written. In the algorithm, we first declare the swarm using Particle class which has following properties :

\begin{description}
\item[pBest]: Best position of the particle where the particle is fittest.
\item[particlePosition] : Particle present position.
\item[particleError] : Particle present error determined by fitness function.
\end{description}

Fitness function in the algorithm computes the value of the mathematical function with the position of the particle, the value is also called error.For each particle, fitness is calculated for every position the particle is in. Our goal here is to find the position where the value returned by the fitness function is minimum. If the present fitness is better than the particle best fitness so far, we will update the particle's best position. Global best position is the best position among all the particles in the swarm. In every iteration, the global best and particle best are updated and all the particles will move closer to the particle that gives global best position. From there each particle moves randomly for a particular distance, this distance is calculated as velocity v in every iteration and depends on learning factors : c1,c2 \cite{Slovik} \cite{Rooy}.\\

\begin{algorithm}[!ht]
\SetAlgoLined
\KwResult{ Optimal Solution for a problem }
  p = Particle()\;
  swarm = [p] * numberOfParticles\;
 \While{ Error approximates to minimum possible value}{

  \For{p in swarm}{
  fp = fitness(particlePosition)\;
  \If{fp is better than fitness(pBest)}{
  pBest = p
  particleError = fp
  }
  }
  gBest = best particlePosition in swarm\;
  gError = best particleError in swarm\;
  \For{particle in swarm}{
  v = v + c1*rand*(pBest - particlePosition) + c2*rand*(gBest - particlePosition)\;

  }
 }
 \caption{Particle Swarm Optimization Algorithm}
\end{algorithm}

\subsection{Bokeh \&\ Panel}

We used Panel, an open-source python library to create interactive visualization and dashboard. The dashboard layout is designed with pn.row, pn.column to place a plot or widget in row \&\ column. Any widget placed in the panel is a container that has a certain functionality and user can utilize them to tweak the parameters of the particle swarm algorithm. Additionally, we also deployed slider widgets using the integer sliders functionality of the panel to choose the number of particles and also facilitated the user with a drop-down box to choose from different mathematical functions. To plot graphs and achieve a continuous streaming of particles we used a holoviews dynamic map container and the coordinates of the particles are updated for every 3 seconds using a periodic callback function. The changes in the widgets are applied to algorithm with slider value to plot accordingly. We have also create buttons when clicked will start the swarm. Any intermediate changes during the streaming is also handled. This entire user-interface is hosted by bokeh server using tornado, where tornado is a asynchronous python networking library.

\subsection{Ray}
Ray enabled us to make distributed computing possible with code changes. We have to initiate the ray using ray.init function to initialize the ray context. A ray.remote decorator upon a function that will be executed as a task in a different process. A .remote post-fix is used to get back the work done at processes of a remote function method. The important concepts to understand in ray are ray nodes and ports, to run the application in a distributed paradigm we start the process of distribution by starting the head node first and later the worker nodes will be given the address of a head node to form a cluster and the ray worker nodes are automatically scalable upon the workload of application. The inter-process communication between each worker process is carried via TCP ports, an additional benefit of using ray nodes is their security.\\

\subsection{Experimental analysis}
The swarm particles visualization plot for the mathematical function : $x^2+(y-100)^2$.\\
\begin{figure}[!ht]
  \centering
  \includegraphics[width=4in]{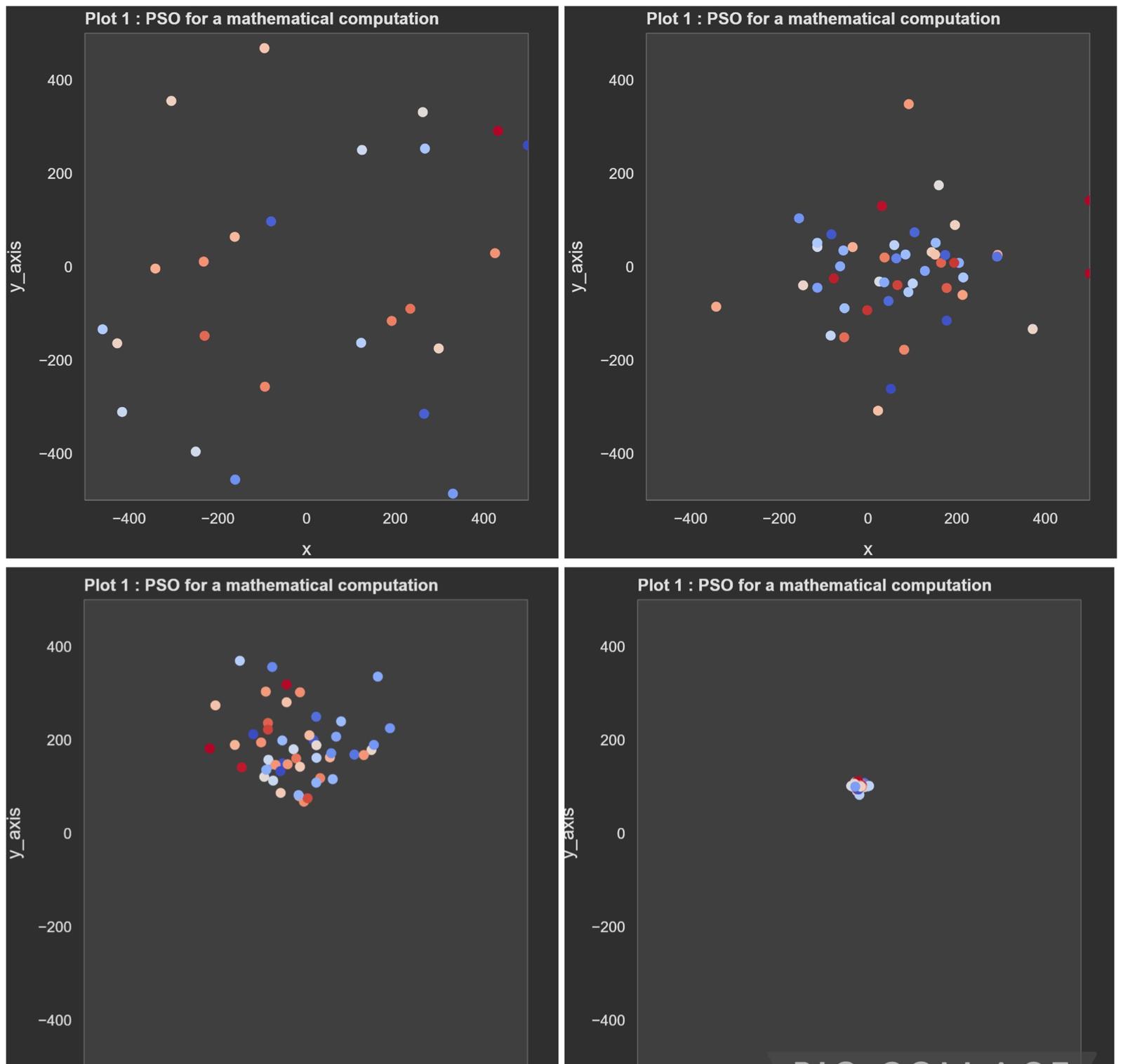}
  \caption{50 Particles solving a mathematical function}
\end{figure}

The swarm particles visualization plot for the mathematical function : $(x-234)^2+(y+100)^2$.\\
\begin{figure}[!ht]
  \centering
  \includegraphics[width=4in]{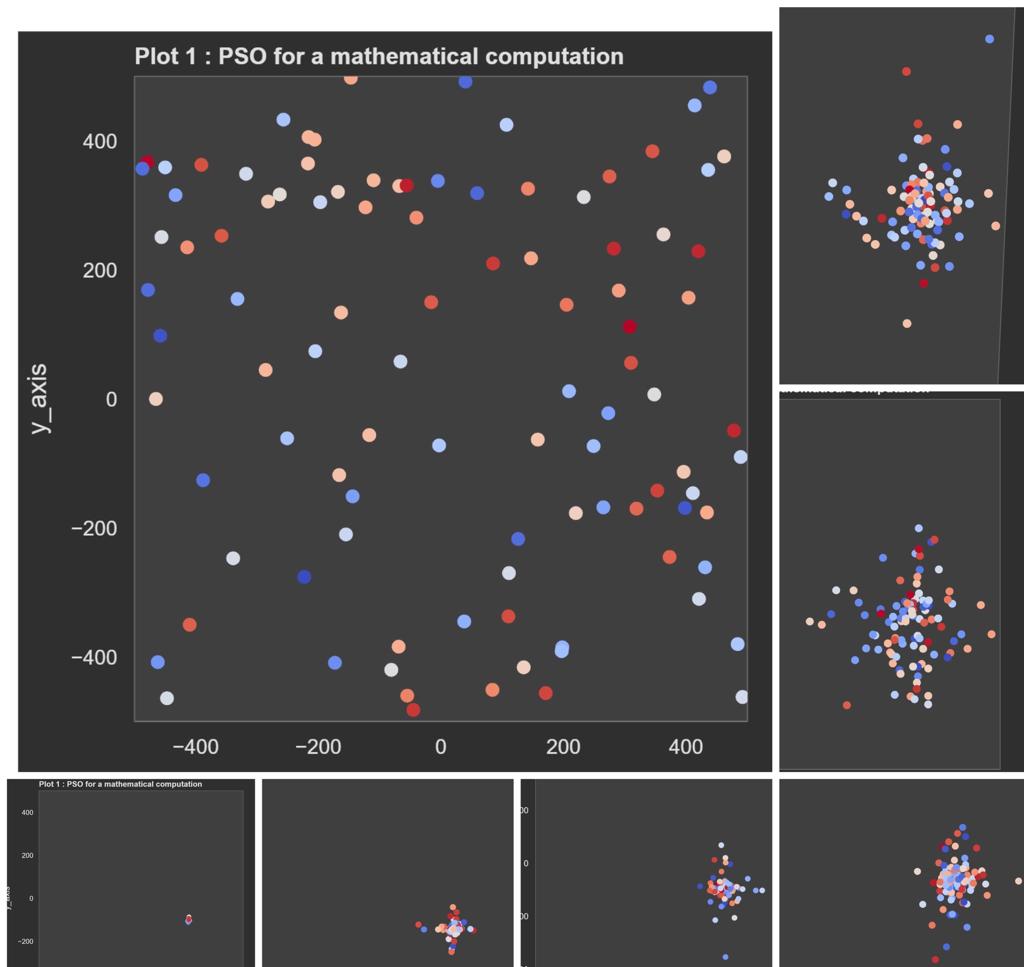}
  \caption{100 Particles solving a mathematical function}
\end{figure}

\section{Conclusion}

A web application for visualising the Particle Swarm Optimization algorithm is implemented with Ray for scalability in this project. The computing process sent to Ray worker nodes has effectively progressed. In our experimental analysis, the system architecture has met all desired distributed challenges. Similarly, the effectiveness of swarm intelligence behaviour is now simple to understand with this application. For future research, we would like to adapt this framework to other optimization problems and evaluate their performance. Also, enable users to input their mathematical function in the dashboard for particles to swarm and give an error plot of their function with PSO.\\

\bibliographystyle{unsrt}  


\begin{thebibliography}{1}

\bibitem{gupta}
Gupta, Sahil. 
\newblock Introduction to swarm intelligence. 
\newblock{\em GeeksforGeeks, (2021, May 15). Retrieved March 5, 2022,} from https://www.geeksforgeeks.org/introduction-to-swarm-intelligence/ 

\bibitem{Kennedy}
Kennedy, J.; Eberhart, R.
\newblock Particle swarm optimization.
\newblock {\em Proceedings of ICNN'95 - International Conference on Neural Networks (1995), 4(0), 1942\(-\)1948}, doi:10.1109/icnn.1995.488968.

\bibitem{Moritz}
Moritz, Philipp, et al. 
\newblock Ray: A Distributed Framework for Emerging AI Applications.
\newblock {\em ArXiv.org, ArXiv, 16 Dec 2017}, arXiv:1712.05889v2. 


\bibitem{lind}
Lindfield, G.; Penny, J. 
\newblock Particle swarm optimization algorithms.
\newblock {\em Introduction to Nature-Inspired Optimization, 18 August 2017}, Retrieved from https://www.sciencedirect.com/science/article/pii/B9780128036365000037. 

\bibitem{rud}
Rudiger, P.
\newblock Panel: A high-level app and dashboarding solution for the PyData ecosystem. \newblock{\em Medium, (2019, June 3).}, https://medium.com/@philipp.jfr/panel-announcement-2107c2b15f52.

\bibitem{shirako}
Shirako, J., Hayashi, A., Paul, S. R., Tumanov, A., \& Sarkar, V. 
\newblock Automatic parallelization of python programs for distributed heterogeneous computing.
\newblock{\em arXiv.org, arXiv, 11 March 2022}, from https://doi.org/10.48550/arXiv.2203.06233. 

\bibitem {phillipp}
Philipp Moritz and Robert Nishihara and Stephanie Wang and Alexey Tumanov and Richard Liaw and Eric Liang and Melih Elibol and Zongheng Yang and William Paul and Michael I. Jordan and Ion Stoica
\newblock Ray: A Distributed Framework for Emerging {AI} Applications.
\newblock{\em inproceedings of 13th USENIX Symposium on Operating Systems Design and Implementation (OSDI 18), October 2018}, isbn 978-1-939133-08-3, Carlsbad, CA,pages 561--577, USENIX Association.

\bibitem{Slovik}
Slovik, Adam. 
\newblock Swarm Intelligence Algorithms: A Tutorial. 
\newblock {\em 1st ed.}, CRC PRESS, 2020. 

\bibitem{Rooy}
Rooy, N. (n.d.).
\newblock{\em Particle swarm optimization from scratch with python}. nathanrooy.github.io. Retrieved from https://nathanrooy.github.io/posts/2016-08-17/simple-particle-swarm-optimization-with-python/ 
\end{thebibliography}

\end{document}